\documentclass[a4paper,conference]{IEEEtran}
\IEEEoverridecommandlockouts
\usepackage{cite}
\usepackage{amsmath,amssymb,amsfonts}
\usepackage{algorithmic}
\usepackage{graphicx}
\usepackage{textcomp}
\usepackage{xcolor}
\usepackage{booktabs}
\usepackage{color}
\usepackage{multirow}
\usepackage{makecell}
\usepackage{stmaryrd}
\usepackage{mathrsfs}
\usepackage{hyperref}

\def\BibTeX{{\rm B\kern-.05em{\sc i\kern-.025em b}\kern-.08em
    T\kern-.1667em\lower.7ex\hbox{E}\kern-.125emX}}
\begin{document}

\title{Classifying Eye-Tracking Data Using Saliency Maps\\
 \thanks{This research was supported by the ICT Division, Ministry of Posts, Telecommunications and Information Technology of the Government of Bangladesh.}
}

\author{\IEEEauthorblockN{Shafin Rahman}
\IEEEauthorblockA{\textit{Electrical \& Computer Engineering} \\
\textit{North South University}\\
shafin.rahman@northsouth.edu}
\and
\IEEEauthorblockN{Sejuti Rahman}
\IEEEauthorblockA{\textit{Robotics \& Mechatronics Engineering} \\
\textit{University of Dhaka}\\
sejuti.rahman@du.ac.bd}
\and
\IEEEauthorblockN{Omar Shahid}
\IEEEauthorblockA{\textit{Robotics \& Mechatronics Engineering} \\
\textit{University of Dhaka}\\
omarshahid232@gmail.com}
\and
\IEEEauthorblockN{Md. Tahmeed Abdullah}
\IEEEauthorblockA{\textit{Robotics \& Mechatronics Engineering} \\
\textit{University of Dhaka}\\
a.tahmeed@yahoo.com}
\and
\IEEEauthorblockN{Jubair Ahmed Sourov}
\IEEEauthorblockA{\textit{Robotics \& Mechatronics Engineering} \\
\textit{University of Dhaka}\\
sourovejubair@gmail.com}
}

\def\edit#1{{\color{orange} {{#1}}}} 

\def\OR#1{{\color{cyan} {{#1}}}} 
\def\TA#1{{\color{blue} {{#1}}}} 
\def\JH#1{{\color{green} {{#1}}}} 
\def\SR#1{{\color{magenta} {{#1}}}} 
\def\SJ#1{{\color{red} {{#1}}}} 


\maketitle

\begin{abstract}
A plethora of research in the literature shows how human eye fixation pattern varies depending on different factors, including genetics, age, social functioning, cognitive functioning, and so on. Analysis of these variations in visual attention has already elicited two potential research avenues: 1) determining the physiological or psychological state of the subject and 2) predicting the tasks associated with the act of viewing from the recorded eye-fixation data. To this end, this paper proposes a visual saliency based novel feature extraction method for automatic and quantitative classification of eye-tracking data, which is applicable to both of the research directions. Instead of directly extracting features from the fixation data, this method employs several well-known computational models of visual attention to predict eye fixation locations as saliency maps. Comparing the saliency amplitudes, similarity and dissimilarity of saliency maps with the corresponding eye fixations maps gives an extra dimension of information which is effectively utilized to generate discriminative features to classify the eye-tracking data. Extensive experimentation using Saliency4ASD \cite{duan_2019_dataset}, Age Prediction \cite{Dalrymple_2019_nature}, and Visual Perceptual Task \cite{koehler_2014_saliency} dataset show that our saliency-based feature can achieve superior performance, outperforming the previous state-of-the-art methods \cite{Chen_2019_ICCV,Dalrymple_2019_nature,boisvert_2016_predicting} by a considerable margin. Moreover, unlike the existing application-specific solutions, our method demonstrates performance improvement across three distinct problems from the real-life domain: Autism Spectrum Disorder screening, toddler age prediction, and human visual perceptual task classification, providing a general paradigm that utilizes the extra-information inherent in saliency maps for a more accurate classification.
\end{abstract}

\begin{IEEEkeywords}
Eye-tracking, Visual Saliency, Autism Spectrum Disorder, Toddler Age Prediction, Visual Perceptual Task
\end{IEEEkeywords}

\section{Introduction}

Eye-tracking is the technological process of recording gaze movements and viewing patterns across time and task. With an eye-tracking camera, one can capture eye movements like fixations, saccades, and smooth pursuit. The lengths/durations of these movements vary with the subconscious state of the brain. Cognitive processes of attention, such as perception, memory, and decision making further influence human gaze behavior \cite{Carter_2020_journal_of_psychophysiology}. As a result, eye-tracking data provides valuable insights for psychology, neuroscience, visual behavior study, visual stimuli response measurement, and so on. As eye-tracking technology is easily available, cheap and minimizes expert manual interventions of the subjective process, nowadays, it has become an integral part of gaze-based command and control \cite{bates_2007_researchgate}, user interface design \cite{yin2018classification}, diagnosing cognitive disorders \cite{ahonniska_2018_EJBN(retts)}.

\begin{figure}[!t]
    \centering
   \includegraphics[width=.5\textwidth,trim={0cm 0cm 0cm 0cm},clip]{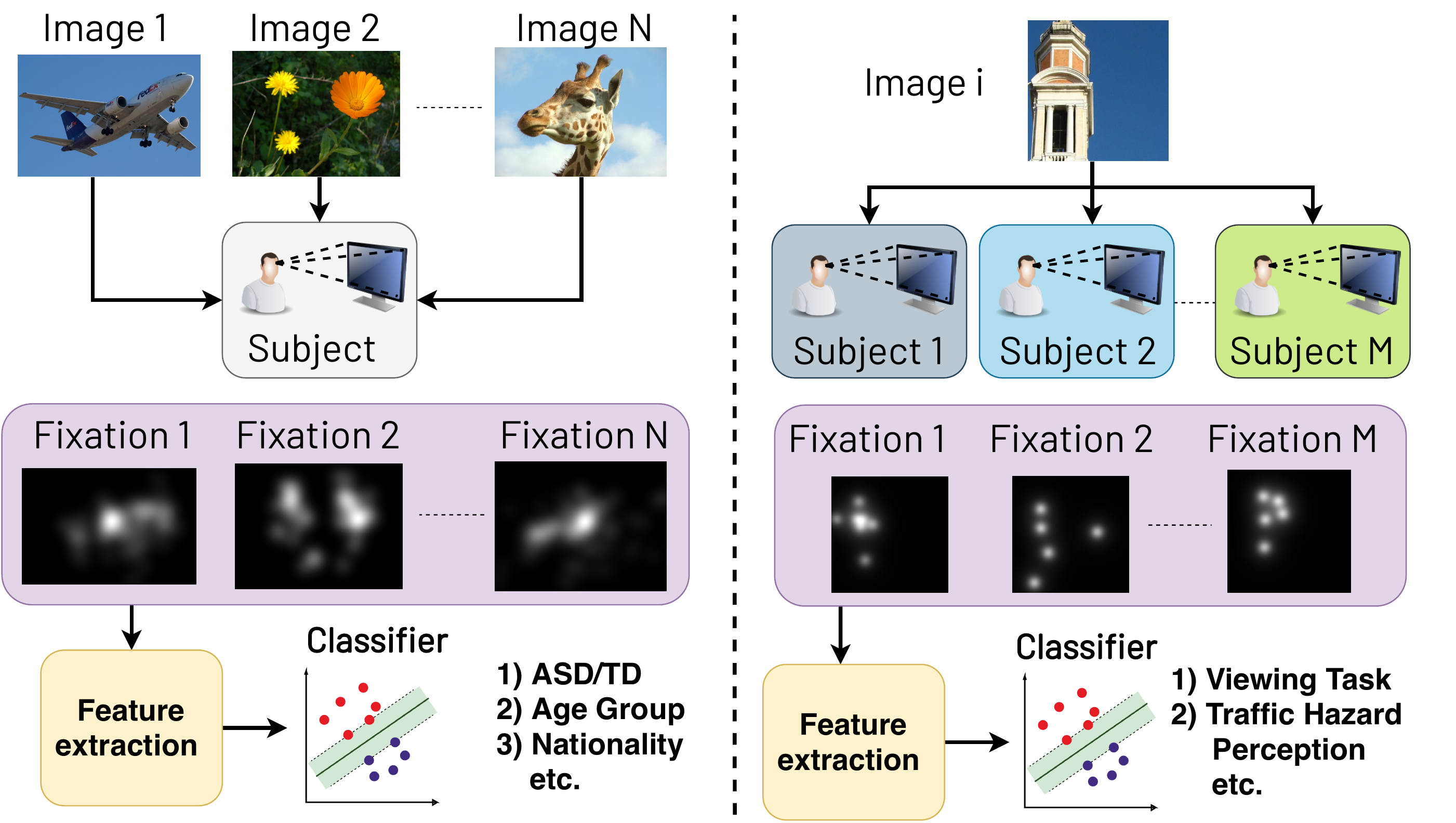}
   \vspace{-1.4em}
    \caption{Problem overview. The classification system of eye-tracking data can be of two types. (a) Subject classification: Given eye-tracking data of one subject while viewing a series of images, the system identifies the category of the subject. E.g., Autism Spectrum Disorder (ASD) screening \cite{Chen_2019_ICCV}, age group classification \cite{Dalrymple_2019_nature}, etc. (b) Visual perception classification: Given eye-tracking data of multiple subjects, while viewing one image, the system identifies the visual perception involved with the image. E.g.,viewing task classification \cite{boisvert_2016_predicting}, traffic hazard perception problem \cite{tafaj_2013_ICANN}, etc. In this paper, we propose a novel feature extraction method that is equally useful across various eye-tracking classification problems.}
   \label{fig:overview}
\end{figure}



Eye-tracking data describes the human visual attention behavior. In general, three factors dictate this behavior. (a) Input stimuli/images like social cues, presence or absence of objects and their locations, (b) observer's age, psychological or physiological state, ethnicity, and (c) tasks associated during the act of viewing like free viewing, object search etc. Therefore, we can obtain important details about image/subject/tasks by classifying eye-tracking data. Prior research in this line of investigation usually targets a single real-life problem, e.g., (i) Autism Spectrum Disorder (ASD) and Typically Developed (TD) children classification \cite{Chen_2019_ICCV}, (ii) toddler age prediction \cite{Dalrymple_2019_nature}, (iii) bio-metric identification \cite{liang2012video}, (iv) behavior analysis of user interface \cite{yin2018classification}, (v) traffic hazard perception, (vi) visual perception of user \cite{boisvert_2016_predicting}, etc. Although the end goals of these problems are different, each of the work aims to address one common task: to classify eye-tracking data while viewing images/videos. In this paper, we attempt to solve the common task so that the same solution can seamlessly work across different real-life problems. We categorize state-of-the-art eye-tracking classification research into two types (see Fig \ref{fig:overview}), (a) subject classification where eye-tracking data includes fixations of each subject for a set of images. The goal is to classify subjects by analyzing their fixations (i - iv).  (b) Visual perception classification where eye-tracking data includes fixation of a group of subjects for an image. The goal is to classify the image type or visual task involved while capturing the fixation data (v - vi). This paper proposes a novel task-agnostic feature extraction method for eye-tracking classification that works irrespective of the problem domain.

For feature extraction, traditional approaches use HoGs \cite{martinez_2012_ICIP}, Gist \cite{li_2009_vision}, Spatial density \cite{kurzhals_2013_VCG}, LM filters \cite{boisvert_2016_predicting}, CNN feature (VGG \cite{jiang_2017_ICCV,nebout_2019_ICMEW}, ResNet \cite{wong_2019_ICPCCW}). We notice that the same feature set does not consistently work across problems. The possible reasons could be (a) different problems require to find different aspects of fixation data as distinguishing information, and (b) the learning model could not get enough supervision from a small amount of fixation data. To get rid of these problems, we employ saliency maps from established saliency models (e.g., GBVS \cite{harel_2007_graph}, CovSal \cite{erdem_2013_visual}, SimpSal \cite{itti_1998_model} etc) in the literature. We compare the input fixation maps with saliency maps using the saliency evaluation metric (e.g., sAUC \cite{borji_2013_ImageProcessing}, CC \cite{le_2007_Visionresearch}, NSS \cite{borji_2013_ImageProcessing}, etc.) and use the evaluation results to construct our feature set. It measures the variation between those two maps, providing useful information to classify eye-tracking data of different problems. Moreover, since saliency maps estimate where people generally look, it adds extra-supervision signals to the learning model about the characteristic of the visual stimuli or image. We apply our feature extraction process in three well-known eye-tracking classification problems: ASD screening, toddler age prediction, and human visual perceptual task classification. We establish a new state-of-the-art performance using Saliency4ASD \cite{duan_2019_dataset}, Age Prediction \cite{Dalrymple_2019_nature}, Visual Perceptual Task \cite{koehler_2014_saliency} datasets.

Our overall contributions are as follows.
\begin{itemize}
    \item We propose a novel feature extraction method that helps to classify eye-tracking data across several real-life problem domains.
    \item We introduce that saliency maps from popular saliency models can be a powerful tool to extract discriminative features for fixation data.
    \item We provide extensive experiments on three eye-tracking datasets and seamlessly achieve state-of-the-art performance on three tasks, e.g., ASD screening, toddler age and visual perceptual task prediction.
\end{itemize}



\section{Related Work} 
Eye-tracking data includes fixation patterns and movement of the observers. In a broad sense, research on the classification of eye-tracking data can be of two types. One is to analyze the raw data and detect distinct type of eye movement events (i.e., fixation \cite{larsson_2015_BSPC}, scan-path \cite{burch_2019_Symposium}, saccades \cite{behrens_2010_Behavioural}) and the other is to classify user groups \cite{Dalrymple_2019_nature,Chen_2019_ICCV}, nationality \cite{yin2018classification} or visual tasks \cite{haji_VR_2014, boisvert_2016_predicting} through analysis of eye-tracking data. This paper mostly follows the second body of works, where we are interested in solving real-life problems with the help of eye-tracking data.

Eye movement data correlates with human psychology, physiology, and many other fields. Researchers analyze the variability in saccade movements for medical diagnosis of neurodegenerative disorders \cite{anderson_2013_nature}, schizophrenia \cite{holzman_1974_JAMA_psychiatri(schizophrenia)}, and acute alcohol consumption \cite{childs_2012_(alcohol)}. In the same vein, the impact of age on visual attention and gaze patterns have been studied widely. Different age groups (i.e., 2, 4-6, 6-8, 8-10 years old) are classified based on gaze data, saliency map agreement, and center bias tendency of subjects \cite{rahman_PLOSONE_2015}. Toddlers of 3 and 30 months are classified based on the differences of their fixations on video stimulus. A data-driven approach using deep learning is adopted to illustrate the factors that drive the inter-individual changes in gaze patterns because of age\cite{Dalrymple_2019_nature}. Given the fixation of subjects on a fixed set of stimuli, they can classify the age of that subject into two classes (18 months and 30 months).

Automatic ASD detection system allows diagnosing individuals with less complexity and without any aid of an expert physician \cite{Shahid_2020_preprint}. The gaze pattern of the image viewing provides discriminative features, which are a powerful tool for classifying ASD. ASD individuals show atypical patterns in face-scanning \cite{liu_2016_identifying} as they show less attention to prime features of faces like eyes, nose, mouth during face-scanning \cite{syeda_2017_visual}. Another study shows ASD and control groups spent a variable amount of fixation time \cite{dris_2019_ICCAIS,wan_2019_ASD} in different regions of interest in an image. Moreover, some studies have found ASD-TD group to show distinct types of scan paths during image viewing \cite{arru_2019_ICMEW,shihab_2020_Bioinformatics} as their cognitive ability and area of interest in an image are distinct from one another. Chen and Zhao \cite{Chen_2019_ICCV} reported a study that used temporal information of eye movements during image viewing, which decodes discriminative features of ASD and healthy group children. Moreover, this study presented a privileged modality framework that utilizes multiple behavioral data sources and provides a better result compared with state-of-the-art performance. In this paper, given eye-tracking data of subject group of interest, we focus on ASD screening and toddler age prediction task.

Apart from the subject group classification problem, eye gaze data can also be used to classify the viewing pattern \cite{haji_VR_2014} or analyze the viewing scene \cite{tafaj2013online} and identify biometric pattern \cite{liang2012video}. In a study, Tafaj \textit{et al.} \cite{tafaj2013online} classify hazardous driving situations based on the driver's eye-tracking data. In another study, \cite{boisvert_2016_predicting} attempted to predict visual tasks from eye movement trajectories of multiple subjects. They studied four different visual tasks: free viewing (observing images without any particular goal), object search (searching a particular object in images), saliency viewing (finding whether the left or right side of images are salient),  explicit judgment (manually selecting the most important location of the image). In this paper, we apply our proposed feature extraction method on the visual task-based classification.

\section{Method}



Researchers have already proposed automated systems to  classify unique characteristics of the image (hazardous situation in driving), observer's class (age, ASD-TD prediction), or visual tasks (free view, explicit judgment) prediction based on fixation (eye-tracking) data. From the perspective of algorithmic design, the input to such systems can be of two types. \textit{Firstly}, for the observer/subject classification case, an algorithm takes fixation data of all images provided by a single subject as input to predict characteristics of the subject under study. \textit{Secondly}, for image or task classification cases, fixation data of all subjects originating from a single image are provided to the algorithm to classify the image or task involved in the data collection process. In line with the discussion above, now, we formally describe eye-tracking classification problems.

\subsection{Problem Formulation}
We assume, $M$ subjects/observers, $\{S_s\}_{s=1}^{M}$, observe $N$ images, $\{I_i\}_{i=1}^{N}$ while providing eye fixation/gaze data. Let, $F_{si}$ denotes fixation data of $s^{th}$ subject for $i^{th}$ image. Suppose, we have $K$ distinct classes, $\{y_c\}_{c=1}^{K}$, that may represent a group (like ASD/TD, age groups, etc.) or viewing activity (like free viewing, saliency viewing etc). The classification of eye-tracking data can be of following two types: 

\begin{itemize}
    \item \textbf{{Subject classification}}: The fixation data of $s^{th}$ subject over $N$ images/trials is $F_s = \{ F_{s1}, F_{s2} \ldots F_{sN} \}$. Given the eye-tracking data, $F'_s$, of a test subject, our goal is to assign a label $\hat{y}_c$ by learning a classifier on the training set, $ \tau = \{ (F_s, y_c) : s \in [1,M] \text{ and } c \in [1,K] \}$. Example: age prediction \cite{Dalrymple_2019_nature}, ASD screening \cite{Chen_2019_ICCV}. 
    
    \item \textbf{{Visual perceptual task classification}}: The fixation data of $i^{th}$ image collected from all $M$ observer/subjects is $F_i = \{ F_{i1}, F_{i2} \ldots F_{iM} \}$. Given the fixation data of test image $F'_i$, our goal is to classify an activity or task label $\hat{y}_c$, by learning a classifier on the training set, $ \tau=\{ (F_i, y_c) : i \in [1,N] \text{ and } c \in [1,K] \}$. Example: perceptual task prediction \cite{boisvert_2016_predicting}.
    
\end{itemize}

\begin{figure}[!t]
    \centering
   \includegraphics[width=.49\textwidth,trim={2cm .5cm 1.8cm .2cm},clip]{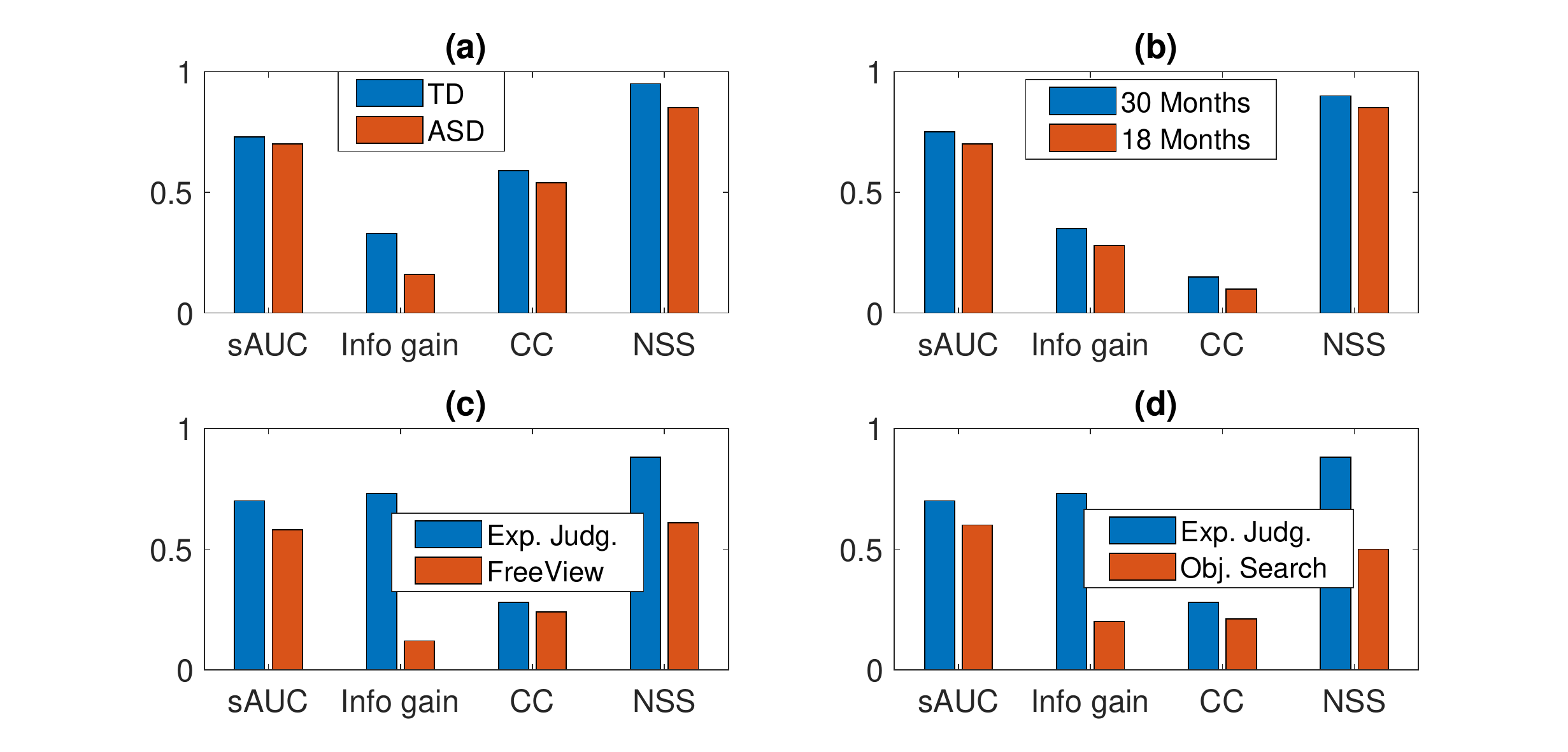}
   \vspace{-2em}
   \caption{Performance (based on sAUC, Info gain, CC, NSS) of saliency model, CIWaM \cite{imamoglu_2012_saliency} while predicting fixation map of (a) ASD/TD, (b) 18/30 months-aged, (c) Free-viewing/Explicit Judgment, and (d) Object Search/Explicit Judgment. One can notice the saliency map predicts TD, 30 months-aged, Saliency-viewing, and Explicit Judgment better than its counter category. Therefore, saliency maps can help to distinguish different types of eye-tracking data.} 
   \label{fig:salmap}
\end{figure}

\begin{figure*}[!t]
    \centering
    \includegraphics[width=1\linewidth]{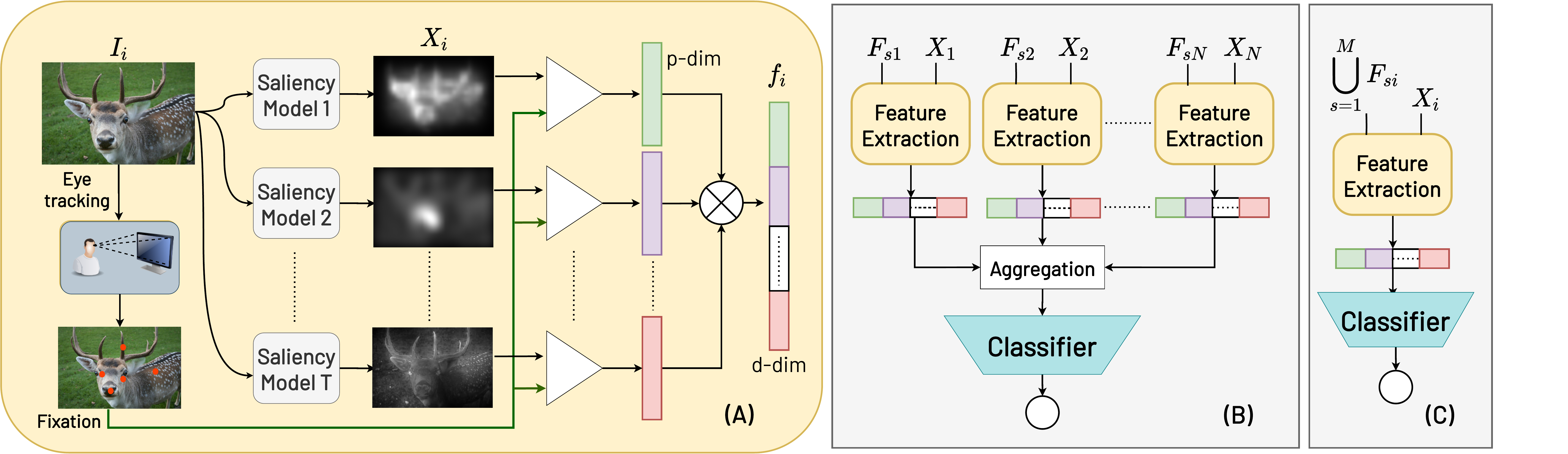}
    \vspace{-1.5em}
    \caption{Our proposed method. \textbf{(A)} Feature extraction process uses $T$ saliency maps (generated using $T$ saliency models) of an image that are compared with the fixation map of that image to generate a $p$ dimensional evaluation vector from each saliency map. The evaluation vectors from $T$ saliency maps are concatenated to get the $d=T \times p$ dimensional vector, which is used for training. Here, $\triangleright$ and $\otimes$ means comparison and concatenation, respectively. \textbf{(B)} For subject group classification, to classify the $sth$ subject, the feature vectors generated from fixation data on $N$ images (presented to that subject) are aggregated to form one $d$-dimensional vector. \textbf{(C)} Training for visual perception task. The fixation data for all $M$ subjects on a single image performing the same perception task is used to form a single fixation map, and a single feature vector is created, which is fed forward to the classifier.}
    \label{fig:ourapproach}
\end{figure*}

\noindent\textbf{Solution overview:} We know that saliency models attempt to simulate human attention mechanisms by predicting eye-tracking data \cite{Bylinskii_2019_PAMI,rahman_NIPS_2015}. However, the extent to which a saliency model predicts fixation data of different classes (subject or task) should differ from one class to another. For example, in ASD/TD case, a TD subject pays special attention to socially relevant cues, e.g., faces and eyes, whereas a subject with autism focuses less on people and faces \cite{Wang_2015_neuron}. Here, a saliency model will predict TD fixation data better than ASD data because TD's behavior is analogous to the common behavior of people. Similarly, in toddler age prediction case, the saliency maps could model the eye-tracking behavior of 30 months toddlers better than 18 months because 30 months of toddlers fixate on foreground objects whereas 18 months old toddlers pay attention only to human faces \cite{Dalrymple_2019_nature}. In the perceptual task prediction case, saliency models predict the explicit judgment data better than free-viewing because free-viewing data is more prone to add noise than the explicit judgment case \cite{rahman_PLOSONE_2015}. In this way, considering saliency maps as a standard prediction and comparing other fixation maps of different classes with this standard, one can distinguish fixation data. In Fig. \ref{fig:salmap}, by comparing with popular saliency evaluation metric, sAUC \cite{borji_2013_ImageProcessing}, Info gain \cite{kummerer_2014_arXiv}, CC \cite{le_2007_Visionresearch} and NSS \cite{borji_2013_ImageProcessing}, we show that a saliency model like CIWaM \cite{imamoglu_2012_saliency} predicts TD, 30 months toddler, explicit judgment data better than ASD, 18 months toddler, free view data. Being motivated by this trend, in this paper, we propose a feature extraction method by employing saliency maps from established saliency models.

\subsection{Proposed Model}

We illustrate our approach in Fig. \ref{fig:ourapproach}. It has two components: feature extraction and learning a classifier. Depending on the problem type mentioned above, the input to the feature extraction process varies slightly. Now, we discuss the components of our method in detail.

\noindent\textbf{Feature Extraction:} We visualize the complete feature extraction process in Fig. \ref{fig:ourapproach}(a). Instead of adopting the traditional way of extracting features directly from the fixation map, we further employ saliency maps from several established saliency models in the feature extraction process. Suppose $T$ is the total number of saliency models used during feature extraction. Let, $X_i = \{X_{i1}, X_{i2}, \ldots X_{iT} \}$ represents the set of saliency maps for the image $I_i$ using all saliency models, where, $X_{il}$ is the saliency map of $l$th saliency model. Our method uses both fixation map $F^*$ and saliency maps $X_i$ to extract a feature vector corresponding to $i$th image. Depending on the problem type, $F^*$ could be either fixation map of a single subject, $F_{si}$ or union of all subjects $\bigcup_{s=1}^{M}F_{si}$. 

To extract features, we evaluate the performance of each saliency model predicting fixation map $F^*$. There are many evaluation methods available in the literature. Each evaluation measures the prediction ability of a saliency model considering different aspects of attention mechanism. For example, sAUC \cite{borji_2013_ImageProcessing} tries to exclude the effect of center bias, a tendency of human fixation to look at the center of the scene. In contract, AUC\_judd \cite{Judd_2012_TechnicalReport} includes center bias because center bias is a natural viewing pattern that should be a part of the evaluation process. With this motivation, we use $p$ number of evaluation metric while comparing $F^*$ and $X_i$. Then, to get a feature vector for $I_i$, we concatenate all results of $p$ evaluation metric corresponds to $T$ saliency models. In this way, the $d (= T \times p)$ dimensional feature vector, $f_i \in \mathbb{R}^{d}$ becomes as follows:

\begin{equation*}
    f_i = [f_{11},f_{12},..,f_{1p}, f_{21},f_{22},..,f_{2p},\ldots, f_{T1},f_{T2},..,f_{Tp}]
\end{equation*}

Here, each element of $f_i$ represents the result of a saliency evaluation metric (sAUC/AUC\_judd/NSS etc).

\noindent\textbf{Learning Classifier:} We forward the extracted features to a learning model. 
However, based on problem type, we may need to process the extracted features further. For subject classification case shown in Fig. \ref{fig:ourapproach}(b), for each $i = 1,2,...,N$, we apply feature extraction process with $F^* = F_{si}$ and $X_i$ separately. Then, we aggregate the individual feature of $N$ input fixation maps by averaging. In this way, we get a feature vector for $s$th subject considering subject-specific fixation map of all images. Using this feature, we train a classifier to classify the subject. For visual perceptual task classification shown in Fig. \ref{fig:ourapproach}(c), we calculate a union of fixation maps from all subjects regarding a single image. In this case, we use $F^* = \bigcup_{s=1}^{M}F_{si}$ and $X_i$ to extract a feature vector for $i$th image. Again, we forward this feature vector to train a classifier to classify the visual perceptual task associated with that image.

\section{Experiment}

\subsection{Setup}

\noindent\textbf{Dataset:} We experiment with our method using three eye-tracking datasets. Here, we briefly describe those datasets. \textbf{(a)} Saliency4ASD: Duan \textit{et al.} \cite{duan_2019_dataset} collected eye movement data from 28 children where half of them were ASD and other half were TD. This dataset consists of fixation data of 300 selected natural scene images from \cite{judd_2009_ICCV}. For each image, they merged all fixation data from ASD and TD subjects separately to produce fixation maps of representative classes. \textbf{(b)} Age Prediction dataset: Dalrymple \textit{et al.} \cite{Dalrymple_2019_nature} experimented the gaze behavior of 18 and 30 months children. This study selected 100 images from the Object and Semantic Images Eye-tracking (OSIE) database \cite{JuanXu_2014_JoV}, which includes both social and non-social scene. Participants observed each image for three seconds while collecting the data. Fixation data of total 22 and 19 subjects of 18 and 30 months respectively are available from the study. This dataset provides fixation points of each subject and image. \textbf{(c)} Visual Perceptual Task dataset: Koehler \textit{et al.} \cite{koehler_2014_saliency} provided a dataset of humans viewing of 800 natural images while performing four visual tasks (free-viewing, object search, saliency view and explicit judgment of salient region). A total of 20 observers performed free-viewing, object search, and saliency search tasks, whereas 100 observers completed the explicit judgment task. The dataset provides fixation coordinates of each observer across different tasks.

\noindent\textbf{Evaluation Process:} Following the literature of eye-tracking data classification \cite{Chen_2019_ICCV,Dalrymple_2019_nature,boisvert_2016_predicting} , we have evaluated and compared our method with the existing methods using accuracy, sensitivity (i.e. true positive rate), specificity (i.e. true negative rate) and Area Under the ROC Curve (AUC).


\noindent\textbf{Implementation Details\footnote{Codes and evaluation are available at: \url{https://github.com/atahmeed/eye-tracking-with-saliency}}} In this paper, we have used some of the established saliency models for feature extraction. We choose bottom-up saliency models that compute local, global features of the input image to predict saliency without requiring any prior training. The reason for such a choice is that our used datasets \cite{duan_2019_dataset,Dalrymple_2019_nature,koehler_2014_saliency} are not large enough to train top-down based saliency models like 
GazeGan\cite{Che_2020_TIP}, EML-Net \cite{Jia_2020_IVC} etc. Thus, the inadequate training of saliency models might impact our feature extraction process. Considering the points mentioned above, in this study, we have used the following saliency models: (i) CovSal \cite{erdem_2013_visual}, (ii) LDS \cite{fang_2016_learning}, (iii) GBVS \cite{harel_2007_graph}, (iv) UHF \cite{tavakoli_2016_bottom}, (v) CIWaM \cite{imamoglu_2012_saliency}, (vi) CEoS \cite{mairon_2014_closer}, (vii) SimpSal \cite{itti_1998_model}. While comparing the similarity/dissimilarity between saliency maps and the fixation data, we use standard saliecny model evaluation metric \cite{Bylinskii_2019_PAMI} e.g.,  AUC\_Borji \cite{borji_2013_ImageProcessing}, AUC\_Judd \cite{Judd_2012_TechnicalReport}, AUC\_Shuffled \cite{borji_2013_ImageProcessing}, Information Gain (IG) \cite{kummerer_2014_arXiv}, Similarity (SIM) \cite{Judd_2012_TechnicalReport}, Pearson’s Correlation Coefficient (CC) \cite{le_2007_Visionresearch}, Kullback-Leibler divergence (KL-Div) \cite{afgani_2008_Symposium}.  To build our classifiers, we have used the Scikit-learn python package. For SVM classifer, we have applied the kernel trick to employ non-linearity in the classification process. We find our best results by tuning our SVM model's parameter using the polynomial kernel and regularization parameter, $C=0.01$. For XGBoost classifier \cite{Chen_2016_ICKDDM}, we have found the best result using Gradient Boost tree booster with a depth of $3$ and $100$ estimators (weak learners).

\begin{table}[!t]
\centering
\renewcommand{\tabcolsep}{4pt}
\caption{Subject/Observer classification.
}
\begin{tabular}{rcccc}
\toprule
\multicolumn{5}{c}{ASD/TD Classification Results} \\ \midrule

& Accuracy  & Sensitivity  & Specificity  & AUC   \\ \hline

Chen'19 (Independent) \cite{Chen_2019_ICCV}  & 89.00  & 86.00  & 93.00  & 92.00  \\
Chen'19 (Full) \cite{Chen_2019_ICCV}   & 93.00  & 93.00  & 93.00  & 98.00  \\  

Ours (SVM)  & 99.50 & 96.70 & 99.30 & 99.50 \\ 
Ours (XGBoost)  & \textbf{99.80} & \textbf{1.00} & \textbf{99.70} & \textbf{99.80} \\\midrule \midrule

\multicolumn{5}{c}{Toddler Age Classification Results} \\ \midrule

& Accuracy  & Sensitivity  & Specificity  & AUC   \\ \hline
Dalrymple'19 \cite{Dalrymple_2019_nature} & \textbf{83.00} & \textbf{90.00} & 81.00 & \textbf{84.00} \\ 
Ours (SVM)  & 75.60 & 78.90 & 72.70 & 75.80 \\ 
Ours (XGBoost)  & \textbf{83.00} & 84.20 & \textbf{81.80} & 83.00 \\ \bottomrule

\end{tabular}
\label{tab:asd}
\end{table}

\begin{table}[!t]
\renewcommand{\tabcolsep}{4pt}
\centering
\caption{Results of Visual Perceptual Task classification. `-' means unavailable results.}
\scalebox{.95}{
\begin{tabular}{@{}rcccccc@{}}
\toprule
& Free/obj & Free/Sal & Free/Exp & Obj/Sal & Obj/Exp & Sal/Exp \\ \midrule

\multicolumn{7}{c}{All images and subjects} \\ \hline
Boisvert'16\cite{boisvert_2016_predicting}& 84.38 & 66.13 & 89.75 & 89.88 & 97.75 & 90.00 \\
Ours (SVM)  & \textbf{86.35} & \textbf{78.57} & 95.33 & \textbf{94.70} & \textbf{97.80} & \textbf{96.20} \\
Ours (XGBoost)   & 84.20 & 74.30 & \textbf{96.50} & 84.25 & 97.70 & 96.10\\\midrule

\multicolumn{7}{c}{50\% images but all subjects} \\ \hline

Boisvert'16 \cite{boisvert_2016_predicting} & 73.41 & 59.59 & - & 71.01 & - & -  \\ 
Ours (SVM)   & \textbf{79.54} & \textbf{71.70} & \textbf{86.21} & 82.31 & \textbf{90.20} & \textbf{91.56}\\
Ours (XGBoost)   & 78.80 & 69.60 & 86.13 & \textbf{82.51} & 88.60 & 90.36\\\midrule

\multicolumn{7}{c}{All images but 50\% subjects} \\ \hline
Boisvert'16\cite{boisvert_2016_predicting}  & 79.98 & 60.16 & - & 77.85 & - & - \\ 
Ours (SVM)   & \textbf{82.30} & \textbf{66.25} & \textbf{78.77} & \textbf{81.33} & \textbf{84.57} & \textbf{83.18} \\
Ours (XGBoost)   & 77.20 & 64.32 & 75.13 & 80.23 & 79.00 & 81.50 \\
\bottomrule
\end{tabular}}
\label{tab:preception_comapre_1}
\end{table}

\subsection{Overall Result}

We apply our method on three well-known eye-tracking based classification problems. In this subsection, we report our experimental results.

\noindent\textbf{Autism Spectrum Disorder (ASD) screening:} Several works suggested that eye-tracking data could work as a blueprint for ASD vs. TD classification. Such an eye-tracking based method can automate the lengthy, manual, time-consuming, and subjective process of this classification. To facilitate research in this area, Saliency4ASD dataset \cite{duan_2019_dataset} provides 300 fixation maps for each group (ASD and TD) totaling $600$ instances. However, the dataset does not provide any individual subject's fixation. Therefore, similar to the work of Chen and Zhao \cite{Chen_2019_ICCV}, we report cross-validation results based on the leave-one-image-out method in Table \ref{tab:asd}. Chen and Zhao \cite{Chen_2019_ICCV} used an end-to-end deep learning approach where it takes images and fixation maps as input, applies a pre-trained Resnet-50 architecture and used a variant of LSTM network for classification. Their approach reported the highest accuracy of 93\% between two modalities of feature extraction. We notice that our approach successfully outperforms \cite{Chen_2019_ICCV} with a large margin. We get the highest accuracy of 99.8\% using XGBoost as a classifier, although the SVM classifier can also beat state-of-the-art results. Our proposed feature extraction method plays a key role in superior performance. The inherent ability of saliency maps for predicting TD fixation better than ASD helps to classify ASD/TD with high confidence. In contrast,  \cite{Chen_2019_ICCV} learned an LSTM model based on a small ($<$ 600) number of fixation maps, which may not be enough to train a large deep learning model.

\noindent\textbf{Toddler Age Prediction:} We can interpret the variability of age using eye-tracking data. Thus, we can classify different age groups analyzing the variation of fixation data.  Here, we predict the toddler age by examining their gaze behavior. This research could help to monitor the growth of toddlers. In this paper, we perform our experiments on the fixation data of 100 images collected and used by Dalrymple'19 \textit{et al.} \cite{Dalrymple_2019_nature} to predict toddlers belonging to two age groups (18 months and 30 months). The original work used two parallel CNNs (for generating features from fixation data) and an SVM for classification using those features. The CNNs are two pre-trained VGG-16 networks that are trained to predict the difference maps of group fixation data from full-scale and half-scale image input. In our method, we create a feature vector for fixation data of each trial of a subject, and by averaging across the trials, we generate a feature vector for classification. The number of train samples is quite low (41 subjects), and as a result, we validate our training in leave-one-subject-out, similar to the work of \cite{Dalrymple_2019_nature} and report the results in Table \ref{tab:asd}. In contrast to the alternative work, even though we require no training to generate the features, we achieved 83\% accuracy, which is similar to their results.

\noindent\textbf{Perceptual tasks Prediction:} Here, we experiment with koehler \textit{et al.} dataset \cite{koehler_2014_saliency}, where fixation maps of four different visual tasks (i.e., free-viewing, object search, saliency search, and explicit judgment) are present. Given a fixation map of any image,  our goal is to predict the visual task performed while capturing the eye-tracking data. For this, we follow the experimental protocol presented in Boisvert and Bruce \cite{boisvert_2016_predicting}, where a set of binary classification problem is designed based on different combinations of visual tasks (i.e. Free viewing/object search, Free viewing/Saliency viewing, Free viewing/Exp. judgment, Object search/Saliency viewing, Object search/Exp. judgment and Saliency viewing/Exp. judgment). Firstly, we perform 10-fold cross-validation using all images and subjects of the dataset. Secondly, we use fixation data of all subjects from 50\% images in training and rest images in testing. In this case, same subjects are used in both training and testing. Finally, we use fixation data of 50\% subjects from all images in training and rest subjects in testing. In this case, same images are used in both training and testing. We outperform the alternative method \cite{boisvert_2016_predicting} (see Table \ref{tab:preception_comapre_1}) in three mentioned experimental setups. The use of saliency maps provides extra/side information in our method, whereas \cite{boisvert_2016_predicting} relies on conventional features (like HOGs, Gist, Spatial density, LM filters) only from the fixation maps.

\subsection{Ablation Study}


We perform ablation studies in two directions. First, we experiment with varying the number of saliency models used during our proposed feature extraction process. Then, we investigate some alternative baseline feature extraction methods in comparison to our proposed approach.

\begin{figure}[!t]
    \centering
    \includegraphics[width=0.5\textwidth,trim={.8cm 0cm 1.2cm .2cm},clip]{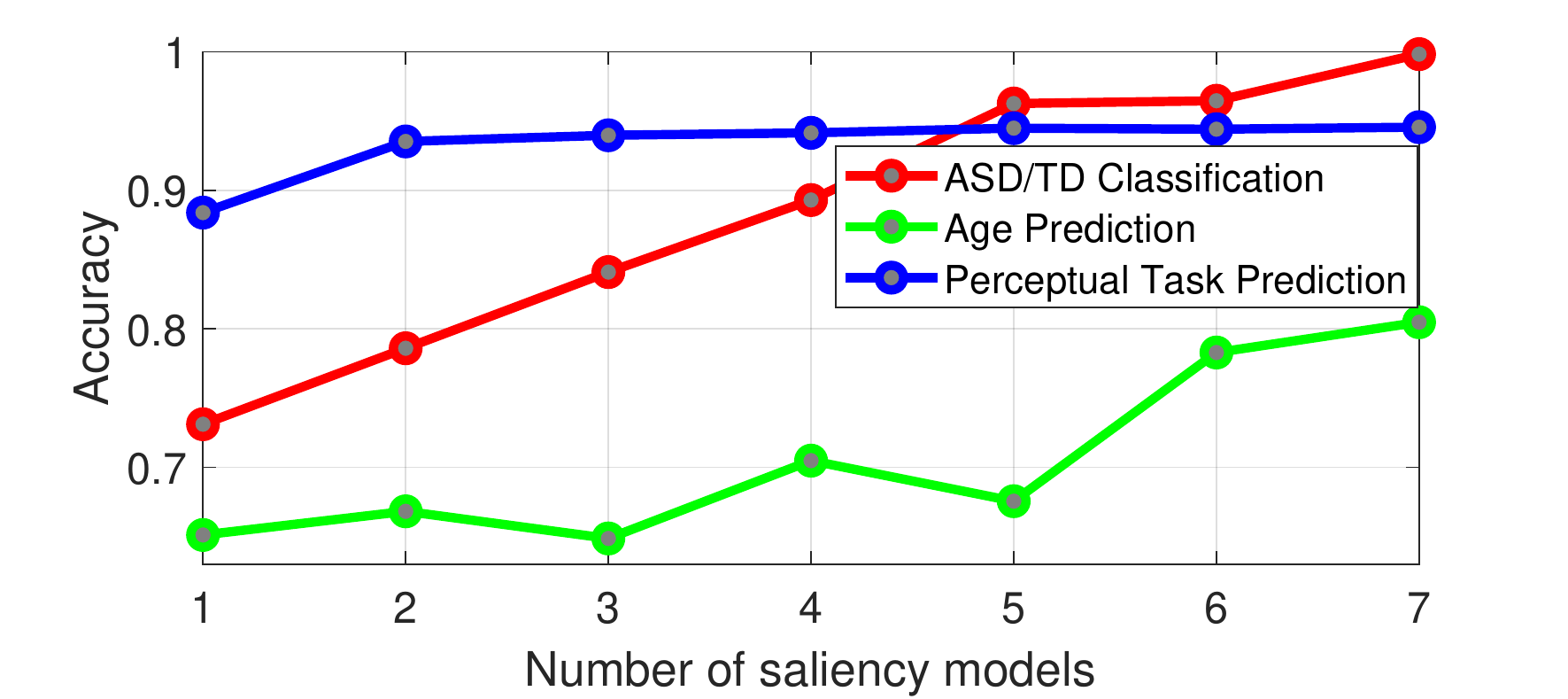}
    \vspace{-2em}
    \caption{Average accuracies of three eye-tracking classification problems using a different number of saliency models during the feature extraction process. One can notice increasing the number of saliency models improves the classification performance up to six saliency models. From six to seven models, the improvement is not significant.}
    \label{fig:asdtd_ablation}
\end{figure}

\noindent\textbf{Varying the number of saliency models:} We perform ablation studies by choosing a different number of saliency models during feature extraction and report classification accuracy on three problems discussed above. Out of seven saliency models used in this paper, we randomly employ 1, 2, 3, 4, 5, 6, and 7 saliency models to perform each problem. Then, we report the average performance by repeating the same experiments ten times. From Fig. \ref{fig:asdtd_ablation} one can notice a clear trend that using more saliency models improves the performance of any problem upto a certain point (using one to six saliency models). After that, the performance becomes stable, providing no significant improvement by adding more saliency models (i.e. adding the seventh model). This trend is expected because after considering a certain amount of saliency models, adding new saliency models cannot add any more discriminative information to the feature extraction process. Thus, the performance becomes stable. 

\noindent\textbf{Baseline methods:} In our approach, we use saliency maps to extract features from fixation. Like traditional approaches, a reasonable alternative can be to obtain features directly from the fixation maps. In this experiment, we apply some popular feature extraction methods, e.g., Histogram of Gradient (HoG), GIST, and VGG-16 on fixation maps and then train XGBoost classifier for eye-tracking classification. This approach does not consider saliency maps in the process. Thus, we consider it as our baseline method. In Table \ref{tab:baseline}, we report the accuracy of the baseline methods and compare the performance with our approach. Our approach successfully outperforms those baselines. It tells that features extracted directly from the fixation maps can not represent the eye-tracking effectively across problems. Such baseline approaches may be useful for some cases like HoG/Gist for perceptual task prediction or VGG16 for age prediction. But, none of the features could work consistently in different problem settings. In contrast, our proposed feature extraction method (using saliency maps) can work seamlessly in multiple real-life problems. Our advantage is that saliency maps indirectly provide extra supervision to the learning process by augmenting distinguishing information about the input visual stimuli.


\begin{table}[!t]
\renewcommand{\tabcolsep}{8pt}
\centering
\caption{Comparison of our method with several baseline feature extraction methods. Baseline methods apply different well-known feature extraction process on the fixation maps. All results are based on the XGBoost classifier.}
\begin{tabular}{@{}cccc@{}}

\toprule
\begin{tabular}[c]{@{}c@{}}Feature\\ Type\end{tabular}       & ASD/TD & Age Prediction & \begin{tabular}[c]{@{}c@{}}Perceptual \\ Task (Free/Obj)\end{tabular} \\ \hline
HoG        &    57.00       &          47.00         &            70.00        \\
Gist       &    68.00      &         57.00          &             81.00       \\
VGG16        &    63.70       &       82.90               &  74.90            \\
Ours       &  \textbf{99.80}     &   \textbf{83.00}            &     \textbf{84.20}               \\ \bottomrule
\end{tabular}
\label{tab:baseline}
\end{table}





\begin{figure*}[!t]
    \centering
    \includegraphics[width=\textwidth,trim={0cm 0cm 0cm 0cm},clip]{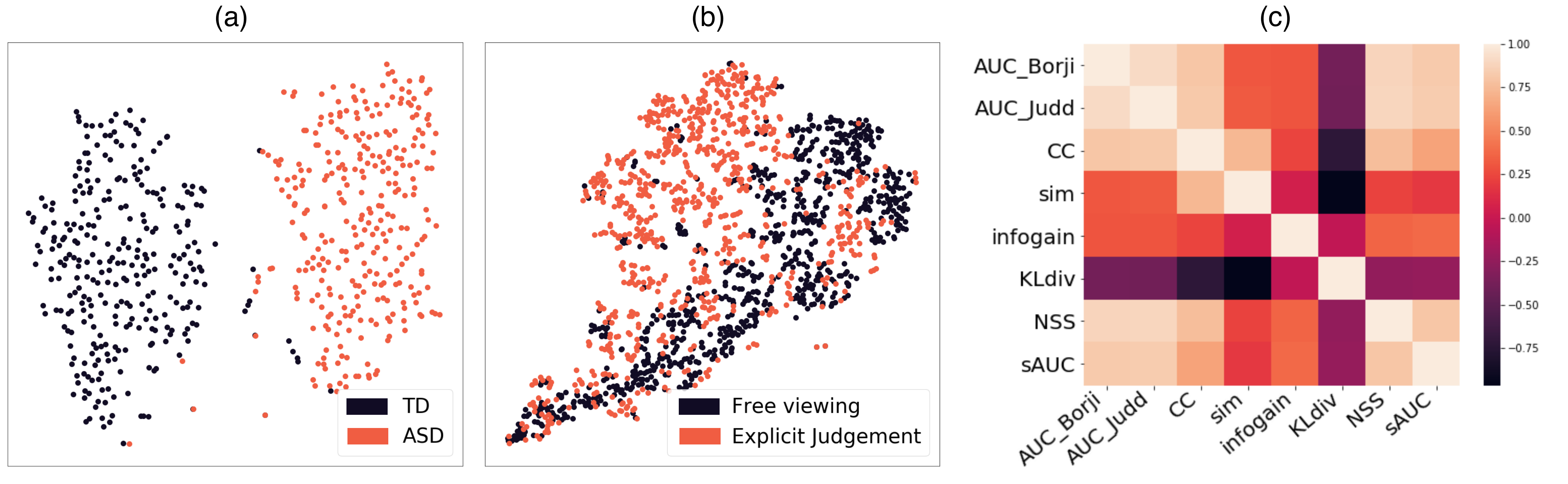}
    \vspace{-2em}
    \caption{2D tSNE \cite{tSNE_van2014} visualization of our extracted features for classification of (a) ASD/TD, (b) Free view/Explicit Judgment. (c) Visualizing the correlation among different saliency evaluation metrices obtained from the perceptual task prediction dataset.}
    \label{fig:tsne}
\end{figure*}
\subsection{Discussion}

In this subsection, we illustrate different critical aspects of our approach and discuss the limitations of this sort of research.

\noindent\textbf{Feature visualization:} We visualize our extracted feature in Fig. \ref{fig:tsne}(a-b). One can notice that the features of different classes are quite well-separated and easy for a simple machine learning model (SVM/XGBoost) for classification. Moreover, we notice that the separation is relatively clear for the case of ASD/TD and Free viewing/Explicit Judgment, where we get better performance (See Table \ref{tab:asd} and \ref{tab:preception_comapre_1}). Furthermore, in Fig. \ref{fig:tsne}(c), we visualize the correlation among different saliency evaluation metric which makes our feature set. The variation of values in non-diagonal regions tells that each evaluation metric focuses on different criteria to compare saliency maps and fixation maps of any particular group. It helps to describe eye-tracking data effectively.

\noindent\textbf{Selection of visual stimuli or social cue:} While classifying a particular subject group (ASD/TD/Toddler Age) or visual perceptual task (Free viewing/Explicit judgment), the images used for capturing the fixation is a paramount concern. For example, subjects with ASD mostly ignore social cues like face, eye, or mouth \cite{liu_2016_identifying}. Toddlers of 18 months of age pay more attention to faces, but 30 months old toddlers focus on foreground objects \cite{Dalrymple_2019_nature}. Explicit judgment fixations mostly concentrate on everyday objects, whereas free viewing fixations become biased towards the center of the image irrespective of object locations \cite{rahman_PLOSONE_2015}. Therefore, the input image set's choice has a profound impact on the performance of the eye-tracking based solution. One can consider designing a process to automatically evaluate how suitable an image is for a particular problem. A simple recommendation could be the Inter-observer congruency (IOC) score \cite{Rahman_ETRA_16}. For a given image, a high IOC score means a high possibility of agreement among common people's viewing patterns. This way, for the same image, if any particular interest group (ASD/TD) views it differently, our feature extraction method will easily pick the distinguishing information to classify that interest group.

\noindent\textbf{Limitations:} A notable limitation of this line of investigation is the unavailability of the large scale and public dataset. All the datasets used in this study (i.e. \cite{duan_2019_dataset,Dalrymple_2019_nature,koehler_2014_saliency}) are small scale containing the number of images and subjects at best 800 and 45, respectively. Moreover, datasets for such problems (ASD screening, Toddler age perdition etc.) are publicly unavailable. In future, one can collect large scale data and open it for researches to further investigate the applicability of eye-tracking data in real-life applications.



\section{Conclusion}
In this paper, we propose a novel feature extraction method for the eye-tracking classification task. The feature extraction process is so generalized that it can solve a wide variety of real-life problems. We employ popular visual saliency models for feature extraction from eye fixation data. Such an approach is better than extracting features from the fixation data alone because saliency maps provide extra supervision to our learning system. We apply our proposed feature extraction process while solving three problems, e.g., ASD screening, toddler age, and visual perceptual task prediction. Our experiments show significant performance boosts in comparison to previous efforts of similar investigations.



\bibliographystyle{ieeetr}
\bibliography{IEEEref}

\begin{thebibliography}{10}

\bibitem{duan_2019_dataset}
H.~Duan, G.~Zhai, X.~Min, Z.~Che, Y.~Fang, X.~Yang, J.~Guti{\'e}rrez, and P.~L.
  Callet, ``A dataset of eye movements for the children with autism spectrum
  disorder,'' in {\em Proceedings of the 10th ACM Multimedia Systems
  Conference}, pp.~255--260, 2019.

\bibitem{Dalrymple_2019_nature}
K.~A. Dalrymple, M.~Jiang, Q.~Zhao, and J.~T. Elison, ``Machine learning
  accurately classifies age of toddlers based on eye tracking,'' {\em
  Scientific Reports}, vol.~9, apr 2019.

\bibitem{koehler_2014_saliency}
K.~Koehler, F.~Guo, S.~Zhang, and M.~P. Eckstein, ``What do saliency models
  predict?,'' {\em Journal of Vision}, vol.~14, no.~3, pp.~14--14, 2014.

\bibitem{Chen_2019_ICCV}
S.~Chen and Q.~Zhao, ``Attention-based autism spectrum disorder screening with
  privileged modality,'' in {\em The IEEE International Conference on Computer
  Vision (ICCV)}, October 2019.

\bibitem{boisvert_2016_predicting}
J.~F. Boisvert and N.~D. Bruce, ``Predicting task from eye movements: On the
  importance of spatial distribution, dynamics, and image features,'' {\em
  Neurocomputing}, vol.~207, pp.~653--668, 2016.

\bibitem{Carter_2020_journal_of_psychophysiology}
B.~T. Carter and S.~G. Luke, ``Best practices in eye tracking research,'' {\em
  International Journal of Psychophysiology}, vol.~155, pp.~49 -- 62, 2020.

\bibitem{bates_2007_researchgate}
R.~Bates, M.~Donegan, H.~O. Istance, J.~P. Hansen, and K.-J. R{\"a}ih{\"a},
  ``Introducing cogain: communication by gaze interaction,'' {\em Universal
  Access in the Information Society}, vol.~6, no.~2, pp.~159--166, 2007.

\bibitem{yin2018classification}
Y.~Yin, C.~Juan, J.~Chakraborty, and M.~P. McGuire, ``Classification of eye
  tracking data using a convolutional neural network,'' in {\em 2018 17th IEEE
  International Conference on Machine Learning and Applications (ICMLA)},
  pp.~530--535, IEEE, 2018.

\bibitem{ahonniska_2018_EJBN(retts)}
J.~Ahonniska-Assa, O.~Polack, E.~Saraf, J.~Wine, T.~Silberg, A.~Nissenkorn, and
  B.~Ben-Zeev, ``Assessing cognitive functioning in females with rett syndrome
  by eye-tracking methodology,'' {\em European Journal of Paediatric
  Neurology}, vol.~22, no.~1, pp.~39--45, 2018.

\bibitem{tafaj_2013_ICANN}
E.~Tafaj, T.~C. K{\"u}bler, G.~Kasneci, W.~Rosenstiel, and M.~Bogdan, ``Online
  classification of eye tracking data for automated analysis of traffic hazard
  perception,'' in {\em International Conference on Artificial Neural
  Networks}, pp.~442--450, Springer, 2013.

\bibitem{liang2012video}
Z.~Liang, F.~Tan, and Z.~Chi, ``Video-based biometric identification using eye
  tracking technique,'' in {\em 2012 IEEE International Conference on Signal
  Processing, Communication and Computing (ICSPCC 2012)}, pp.~728--733, IEEE,
  2012.

\bibitem{martinez_2012_ICIP}
F.~Martinez, A.~Carbone, and E.~Pissaloux, ``Gaze estimation using local
  features and non-linear regression,'' in {\em 2012 19th IEEE International
  Conference on Image Processing}, pp.~1961--1964, IEEE, 2012.

\bibitem{li_2009_vision}
Z.~Li and L.~Itti, ``Gist based top-down templates for gaze prediction,'' {\em
  Journal of Vision}, vol.~9, no.~8, pp.~202--202, 2009.

\bibitem{kurzhals_2013_VCG}
K.~Kurzhals and D.~Weiskopf, ``Space-time visual analytics of eye-tracking data
  for dynamic stimuli,'' {\em IEEE Transactions on Visualization and Computer
  Graphics}, vol.~19, no.~12, pp.~2129--2138, 2013.

\bibitem{jiang_2017_ICCV}
M.~Jiang and Q.~Zhao, ``Learning visual attention to identify people with
  autism spectrum disorder,'' in {\em Proceedings of the IEEE International
  Conference on Computer Vision}, pp.~3267--3276, 2017.

\bibitem{nebout_2019_ICMEW}
A.~Nebout, W.~Wei, Z.~Liu, L.~Huang, and O.~Le~Meur, ``Predicting saliency maps
  for asd people,'' in {\em 2019 IEEE International Conference on Multimedia \&
  Expo Workshops (ICMEW)}, pp.~629--632, IEEE, 2019.

\bibitem{wong_2019_ICPCCW}
E.~T. Wong, S.~Yean, Q.~Hu, B.~S. Lee, J.~Liu, and R.~Deepu, ``Gaze estimation
  using residual neural network,'' in {\em 2019 IEEE International Conference
  on Pervasive Computing and Communications Workshops (PerCom Workshops)},
  pp.~411--414, IEEE, 2019.

\bibitem{harel_2007_graph}
J.~Harel, C.~Koch, and P.~Perona, ``Graph-based visual saliency,'' in {\em
  Advances in Neural Information Processing Systems}, pp.~545--552, 2007.

\bibitem{erdem_2013_visual}
E.~Erdem and A.~Erdem, ``Visual saliency estimation by nonlinearly integrating
  features using region covariances,'' {\em Journal of Vision}, vol.~13, no.~4,
  pp.~11--11, 2013.

\bibitem{itti_1998_model}
L.~Itti, C.~Koch, and E.~Niebur, ``A model of saliency-based visual attention
  for rapid scene analysis,'' {\em IEEE Transactions on Pattern Analysis and
  Machine Intelligence}, vol.~20, no.~11, pp.~1254--1259, 1998.

\bibitem{borji_2013_ImageProcessing}
A.~Borji, D.~N. Sihite, and L.~Itti, ``Quantitative analysis of human-model
  agreement in visual saliency modeling: A comparative study,'' {\em IEEE
  Transactions on Image Processing}, vol.~22, no.~1, pp.~55--69, 2013.

\bibitem{le_2007_Visionresearch}
O.~Le~Meur, P.~Le~Callet, and D.~Barba, ``Predicting visual fixations on video
  based on low-level visual features,'' {\em Vision Research}, vol.~47, no.~19,
  pp.~2483--2498, 2007.

\bibitem{larsson_2015_BSPC}
L.~Larsson, M.~Nystr{\"o}m, R.~Andersson, and M.~Stridh, ``Detection of
  fixations and smooth pursuit movements in high-speed eye-tracking data,''
  {\em Biomedical Signal Processing and Control}, vol.~18, pp.~145--152, 2015.

\bibitem{burch_2019_Symposium}
M.~Burch, A.~Kumar, K.~Mueller, T.~Kervezee, W.~Nuijten, R.~Oostenbach,
  L.~Peeters, and G.~Smit, ``Finding the outliers in scanpath data,'' in {\em
  Proceedings of the 11th ACM Symposium on Eye Tracking Research \&
  Applications}, pp.~1--5, 2019.

\bibitem{behrens_2010_Behavioural}
F.~Behrens, M.~MacKeben, and W.~Schr{\"o}der-Preikschat, ``An improved
  algorithm for automatic detection of saccades in eye movement data and for
  calculating saccade parameters,'' {\em Behavior Research Methods}, vol.~42,
  no.~3, pp.~701--708, 2010.

\bibitem{haji_VR_2014}
A.~Haji-Abolhassani and J.~J. Clark, ``An inverse yarbus process: Predicting
  observers’ task from eye movement patterns,'' {\em Vision Research},
  vol.~103, pp.~127--142, 2014.

\bibitem{anderson_2013_nature}
T.~J. Anderson and M.~R. MacAskill, ``Eye movements in patients with
  neurodegenerative disorders,'' {\em Nature Reviews Neurology}, vol.~9, no.~2,
  pp.~74--85, 2013.

\bibitem{holzman_1974_JAMA_psychiatri(schizophrenia)}
P.~S. Holzman, L.~R. Proctor, D.~L. Levy, N.~J. Yasillo, H.~Y. Meltzer, and
  S.~W. Hurt, ``Eye-tracking dysfunctions in schizophrenic patients and their
  relatives,'' {\em Archives of General Psychiatry}, vol.~31, no.~2,
  pp.~143--151, 1974.

\bibitem{childs_2012_(alcohol)}
E.~Childs, D.~J. Roche, A.~C. King, and H.~de~Wit, ``Varenicline potentiates
  alcohol-induced negative subjective responses and offsets impaired eye
  movements,'' {\em Alcoholism: Clinical and Experimental Research}, vol.~36,
  no.~5, pp.~906--914, 2012.

\bibitem{rahman_PLOSONE_2015}
S.~Rahman and N.~Bruce, ``Visual saliency prediction and evaluation across
  different perceptual tasks,'' {\em PloS One}, vol.~10, no.~9, p.~e0138053,
  2015.

\bibitem{Shahid_2020_preprint}
O.~Shahid, S.~Rahman, S.~F. Ahmed, M.~A. Arrafi, and M.~Ahad, ``Data-driven
  automated detection of autism spectrum disorder using activity analysis: A
  review.,'' {\em Preprints 2020, 2020100388}, 2020.

\bibitem{liu_2016_identifying}
W.~Liu, M.~Li, and L.~Yi, ``Identifying children with autism spectrum disorder
  based on their face processing abnormality: A machine learning framework,''
  {\em Autism Research}, vol.~9, no.~8, pp.~888--898, 2016.

\bibitem{syeda_2017_visual}
U.~H. Syeda, Z.~Zafar, Z.~Z. Islam, S.~M. Tazwar, M.~J. Rasna, K.~Kise, and
  M.~A.~R. Ahad, ``Visual face scanning and emotion perception analysis between
  autistic and typically developing children,'' in {\em Proceedings of the 2017
  ACM International Joint Conference on Pervasive and Ubiquitous Computing and
  Proceedings of the 2017 ACM International Symposium on Wearable computers},
  pp.~844--853, 2017.

\bibitem{dris_2019_ICCAIS}
A.~B. Dris, A.~Alsalman, A.~Al-Wabil, and M.~Aldosari, ``Intelligent gaze-based
  screening system for autism,'' in {\em 2019 2nd International Conference on
  Computer Applications \& Information Security (ICCAIS)}, pp.~1--5, IEEE,
  2019.

\bibitem{wan_2019_ASD}
G.~Wan, X.~Kong, B.~Sun, S.~Yu, Y.~Tu, J.~Park, C.~Lang, M.~Koh, Z.~Wei,
  Z.~Feng, {\em et~al.}, ``Applying eye tracking to identify autism spectrum
  disorder in children,'' {\em Journal of Autism and Developmental Disorders},
  vol.~49, no.~1, pp.~209--215, 2019.

\bibitem{arru_2019_ICMEW}
G.~Arru, P.~Mazumdar, and F.~Battisti, ``Exploiting visual behaviour for autism
  spectrum disorder identification,'' in {\em 2019 IEEE International
  Conference on Multimedia \& Expo Workshops (ICMEW)}, pp.~637--640, IEEE,
  2019.

\bibitem{shihab_2020_Bioinformatics}
A.~I. Shihab, F.~A. Dawood, and A.~H. Kashmar, ``Data analysis and
  classification of autism spectrum disorder using principal component
  analysis,'' {\em Advances in Bioinformatics}, vol.~2020, 2020.

\bibitem{tafaj2013online}
E.~Tafaj, T.~C. K{\"u}bler, G.~Kasneci, W.~Rosenstiel, and M.~Bogdan, ``Online
  classification of eye tracking data for automated analysis of traffic hazard
  perception,'' in {\em International Conference on Artificial Neural
  Networks}, pp.~442--450, Springer, 2013.

\bibitem{imamoglu_2012_saliency}
N.~Imamoglu, W.~Lin, and Y.~Fang, ``A saliency detection model using low-level
  features based on wavelet transform,'' {\em IEEE Transactions on Multimedia},
  vol.~15, no.~1, pp.~96--105, 2012.

\bibitem{Bylinskii_2019_PAMI}
Z.~Bylinskii, T.~Judd, A.~Oliva, A.~Torralba, and F.~Durand, ``What do
  different evaluation metrics tell us about saliency models?,'' {\em {IEEE}
  Transactions on Pattern Analysis and Machine Intelligence}, vol.~41,
  pp.~740--757, mar 2019.

\bibitem{rahman_NIPS_2015}
S.~Rahman and N.~Bruce, ``Saliency, scale and information: Towards a unifying
  theory,'' in {\em Advances in Neural Information Processing Systems 28}
  (C.~Cortes, N.~D. Lawrence, D.~D. Lee, M.~Sugiyama, and R.~Garnett, eds.),
  pp.~2188--2196, Curran Associates, Inc., 2015.

\bibitem{Wang_2015_neuron}
S.~Wang, M.~Jiang, X.~M. Duchesne, E.~A. Laugeson, D.~P. Kennedy, R.~Adolphs,
  and Q.~Zhao, ``Atypical visual saliency in autism spectrum disorder
  quantified through model-based eye tracking,'' {\em Neuron}, vol.~88,
  pp.~604--616, nov 2015.

\bibitem{kummerer_2014_arXiv}
M.~K{\"u}mmerer, T.~Wallis, and M.~Bethge, ``How close are we to understanding
  image-based saliency?,'' {\em arXiv preprint arXiv:1409.7686}, 2014.

\bibitem{Judd_2012_TechnicalReport}
T.~Judd, F.~Durand, and A.~Torralba, ``A benchmark of computational models of
  saliency to predict human fixations,'' in {\em MIT Technical Report}, 2012.

\bibitem{judd_2009_ICCV}
T.~Judd, K.~Ehinger, F.~Durand, and A.~Torralba, ``Learning to predict where
  humans look,'' in {\em 2009 IEEE 12th International Conference on Computer
  Vision}, pp.~2106--2113, IEEE, 2009.

\bibitem{JuanXu_2014_JoV}
J.~Xu, M.~Jiang, S.~Wang, M.~S. Kankanhalli, and Q.~Zhao, ``{Predicting human
  gaze beyond pixels},'' {\em Journal of Vision}, vol.~14, pp.~28--28, 01 2014.

\bibitem{Che_2020_TIP}
Z.~Che, A.~Borji, G.~Zhai, X.~Min, G.~Guo, and P.~L. Callet, ``How is gaze
  influenced by image transformations? dataset and model,'' {\em {IEEE}
  Transactions on Image Processing}, vol.~29, pp.~2287--2300, 2020.

\bibitem{Jia_2020_IVC}
S.~Jia and N.~D. Bruce, ``{EML}-{NET}: An expandable multi-layer {NETwork} for
  saliency prediction,'' {\em Image and Vision Computing}, vol.~95, p.~103887,
  mar 2020.

\bibitem{fang_2016_learning}
S.~Fang, J.~Li, Y.~Tian, T.~Huang, and X.~Chen, ``Learning discriminative
  subspaces on random contrasts for image saliency analysis,'' {\em IEEE
  Transactions on Neural Networks and Learning Systems}, vol.~28, no.~5,
  pp.~1095--1108, 2016.

\bibitem{tavakoli_2016_bottom}
H.~R. Tavakoli and J.~Laaksonen, ``Bottom-up fixation prediction using
  unsupervised hierarchical models,'' in {\em Asian Conference on Computer
  Vision}, pp.~287--302, Springer, 2016.

\bibitem{mairon_2014_closer}
R.~Mairon and O.~Ben-Shahar, ``A closer look at context: From coxels to the
  contextual emergence of object saliency,'' in {\em European Conference on
  Computer Vision}, pp.~708--724, Springer, 2014.

\bibitem{afgani_2008_Symposium}
M.~Afgani, S.~Sinanovic, and H.~Haas, ``Anomaly detection using the
  kullback-leibler divergence metric,'' in {\em 2008 First International
  Symposium on Applied Sciences on Biomedical and Communication Technologies},
  pp.~1--5, IEEE, 2008.

\bibitem{Chen_2016_ICKDDM}
T.~Chen and C.~Guestrin, ``{XGBoost},'' in {\em Proceedings of the 22nd {ACM}
  {SIGKDD} International Conference on Knowledge Discovery and Data Mining},
  {ACM}, aug 2016.

\bibitem{tSNE_van2014}
L.~Van Der~Maaten, ``Accelerating t-sne using tree-based algorithms.,'' {\em
  Journal of Machine Learning Research}, vol.~15, no.~1, pp.~3221--3245, 2014.

\bibitem{Rahman_ETRA_16}
S.~Rahman and N.~D.~B. Bruce, ``Factors underlying inter-observer agreement in
  gaze patterns: Predictive modelling and analysis,'' in {\em Proceedings of
  the Ninth Biennial ACM Symposium on Eye Tracking Research \& Applications},
  ETRA ’16, (New York, NY, USA), p.~155–162, Association for Computing
  Machinery, 2016.

\end{thebibliography}

\end{document}